\documentclass[lettersize,journal]{IEEEtran}
\usepackage{amsmath,amsfonts}
\usepackage{algorithmic}
\usepackage{algorithm}
\usepackage{array}
\usepackage[caption=false,font=normalsize,labelfont=sf,textfont=sf]{subfig}
\usepackage{textcomp}
\usepackage{stfloats}
\usepackage{url}
\usepackage{verbatim}
\usepackage{graphicx}
\usepackage{cite}
\usepackage{hyperref}

\usepackage{amsmath}

\usepackage{amssymb}
\usepackage{multirow}
\newcommand{\upperRomannumeral}[1]{\uppercase\expandafter{\romannumeral#1}}

\usepackage{csquotes}

\usepackage{bbm}
\usepackage{booktabs} % for professional tables
\usepackage{setspace}
\usepackage{adjustbox}
\usepackage{multirow}

\usepackage{amsmath}

\usepackage{amssymb}
\usepackage{pifont}

\usepackage[dvipsnames]{xcolor}

\begin{document}

\title{MatchXML: An Efficient  Text-label Matching Framework for Extreme Multi-label Text Classification}


% \author{Penghang~Liu,
%         Valerio~Guarrasi,
%         and~Ahmet~Erdem~Sar{\i}y\"{u}ce% 
% \IEEEcompsocitemizethanks{\IEEEcompsocthanksitem Penghang Liu and Ahmet~Erdem~Sar{\i}y\"{u}ce are with the Department
% of Computer Science and Engineering, University at Buffalo.\protect\\
% E-mail: \{penghang, erdem\}@buffalo.edu
% \IEEEcompsocthanksitem Valerio~Guarrasi is with the Sapienza University of Rome.\protect\\
% E-mail: valerio.guarrasi@uniroma1.it}% 
% }

% \thanks{Manuscript is under review. Hui Ye is the corresponding author.}

\author
{
 Hui~Ye, Rajshekhar~Sunderraman, and Shihao~Ji, Senior Member, IEEE

\thanks{This work was supported by the Presidential Fellowship in the Transcultural Conflict and Violence Initiative (TCV) at Georgia State University, in part by the National Science Foundation Major Research Instrumentation (MRI) grant number CNS-1920024.}

\IEEEcompsocitemizethanks{
  \IEEEcompsocthanksitem H. Ye, R. Sunderraman, and S. Ji are with the Department of Computer Science at Georgia State University, Atlanta, GA, USA (E-mail: hye2@student.gsu.edu, rsunderraman@gsu.edu, sji@gsu.edu).
 % \IEEEcompsocthanksitem E-mail: hye2@student.gsu.edu, rsunderraman@gsu.edu, sji@gsu.edu
% \IEEEcompsocthanksitem Q. V. H. Nguyen is with the School of Information and Communication Technology, Griffith University.\protect\\
% E-mail: quocviethung.nguyen@griffith.edu.au
% \IEEEcompsocthanksitem K. Zheng is with the Big Data Research Center, University of Electronic Science and Technology of China.\protect\\
% E-mail: zhengkai@uestc.edu.cn
% \IEEEcompsocthanksitem X. Zhang is with the Machine Intelligence and Knowledge Engineering Laboratory, King Abdullah University of Science and Technology.\protect\\
% E-mail: xiangliang.zhang@kaust.edu.sa
% \IEEEcompsocthanksitem H. Wang is with Alibaba AI Labs. E-mail: cashenry@126.com
}

% \thanks{Manuscript is under review. Hui Ye is the corresponding author.}
}



\maketitle

\begin{abstract}

The eXtreme Multi-label text Classification (XMC) refers to training a classifier that assigns a text sample with relevant labels from an extremely large-scale label set (e.g., millions of labels). We propose MatchXML, an efficient text-label matching framework for XMC. We observe that the label embeddings generated from the sparse Term Frequency-Inverse Document Frequency (TF–IDF) features have several limitations. We thus propose \emph{label2vec} to effectively train the semantic dense label embeddings by the Skip-gram model. The dense label embeddings are then used to build a Hierarchical Label Tree by clustering. In fine-tuning the pre-trained encoder Transformer, we formulate the multi-label text classification as a text-label matching problem in a bipartite graph. We then extract the dense text representations from the fine-tuned Transformer. Besides the fine-tuned dense text embeddings, we also extract the static dense sentence embeddings from a pre-trained Sentence Transformer. Finally, a linear ranker is trained by utilizing the sparse TF–IDF features, the fine-tuned dense text representations, and static dense sentence features. Experimental results demonstrate that MatchXML achieves the state-of-the-art accuracies on five out of six datasets. As for the training speed, MatchXML outperforms the competing methods on all the six datasets. Our source code is publicly available at \url{ https://github.com/huiyegit/MatchXML}.

% The eXtreme Multi-label text Classification (XMC) refers to training a classifier that assigns a text sample with relevant labels from an extremely large-scale label set (e.g., millions of labels). To achieve a high performance XMC in terms of accuracy and speed, we propose MatchXML, an efficient text-label matching framework for XMC. We observe that the label embeddings generated from the sparse Term Frequency-Inverse Document Frequency (TF–IDF) features have several limitations. We thus propose \emph{label2vec} to effectively train the semantic dense label embeddings by the Skip-gram model. The dense label embeddings are then used to build a Hierarchical Label Tree by clustering. In the fine-tuning stage of the Transformer-based text encoder, we formulate the multi-label text classification as a text-label matching problem in a bipartite graph. We then extract the dense text representations from the fine-tuned Transformer to boost the model performance. Besides the dense text embeddings fine-tuned from the pre-trained model, we also extract the static dense sentence embeddings from a pre-trained Sentence Transformer. Finally, a linear ranker is trained by utilizing the sparse TF–IDF features, the fine-tuned dense text representations and static dense sentence features. Experimental results demonstrate that MatchXML achieves state-of-the-art accuracy on five out of six datasets. As for the speed, MatchXML outperforms the competing methods on all the six datasets. Our source code is publicly available at \url{ https://github.com/huiyegit/MatchXML}.

\end{abstract}

\begin{IEEEkeywords}
Extreme multi-label classification, label2vec, text-label matching, bipartite graph, contrastive learning
\end{IEEEkeywords}

\section{Introduction}
\IEEEPARstart{T}{he} eXtreme Multi-label text Classification (XMC) refers to learning a classifier that can annotate an input text with the most relevant labels from an extremely large-scale label set (e.g., millions of labels). This problem has many real world applications, such as labeling a Wikipedia page with relevant tags~\cite{dekel2010multiclass}, providing a customer query with related products in product search~\cite{prabhu2018parabel}, and recommending relevant items to a customer in recommendation systems~\cite{covington2016deep}. 

To address the issue of the extremely large output space in XMC, the Hierarchical Label Tree (HLT)~\cite{ prabhu2018parabel} has been proposed to effectively reduce the computational complexity from $O(L)$ to $O(logL)$, where $L$ is the number of labels. Taking label embeddings as input, an HLT can be constructed by partition algorithms~\cite{prabhu2018parabel, yu2022pecos} based on the K-means clustering. Prior works~\cite{prabhu2018parabel, yu2022pecos,chang2020taming,jiang2021lightxml,zhang2021fast} have applied the Positive Instance Feature Aggregation (PIFA) to compute label embeddings, where one label embedding is the summation of the TF–IDF features of the text samples when the label is positive. However, the label embeddings generated from PIFA have several limitations. First, current machine learning algorithms are more efficient to process the data of small dense vectors than the large sparse vectors. Second, the TF–IDF features of text data, which are required by PIFA to generate the label embeddings, may not be always available and thus limits the applications of PIFA. Inspired by the \emph{word2vec}~\cite{mikolov2013distributed, mikolov2013efficient} in training word embeddings, we propose \emph{label2vec} to learn the semantic dense label embeddings. We consider a set of labels assigned to a text sample as an unordered sequence, where each label can be treated as one word/token, and the Skip-gram model~\cite{mikolov2013distributed, mikolov2013efficient} is applied to train the embedding for each label. The \emph{label2vec} approach has better generalization than PIFA as it does not require the TF–IDF features. Besides, the dense label embeddings have smaller storage size that are more efficient to process by the downstream machine learning algorithms. Our experiments demonstrate that the dense label embeddings can capture the semantic label relationships and generate improved HLTs compared to the sparse label embeddings, leading to improved performance in the downstream XMC tasks.

Most of the early works in XMC~\cite{prabhu2014fastxml,bhatia2015sparse,jain2016extreme,yen2016pd,yen2017ppdsparse,babbar2017dismec, tagami2017annexml, prabhu2018parabel, siblini2018craftml,wydmuch2018no,jain2019slice,khandagale2019bonsai, dahiya2021siamesexml, dahiya2021deepxml} leverage the statistical Bag-Of-Words (BOW) or Term Frequency-Inverse Document Frequency (TF–IDF) features as the text representations to train a text classifier. This type of text features is simple, but it can not capture the semantic meaning of text corpora due to the ignorance of word order. Recent works~\cite{liu2017deep,you2019attentionxml,jiang2021lightxml, ye2020pretrained, chang2020taming} explore deep learning approaches to learn the dense vectors as the text representations. These methods leverage the contextual information of words in text corpora to extract the dense text representations, leading to improved classification accuracies. On the other hand, the recently proposed XR-Transformer~\cite{zhang2021fast} and CascadeXML~\cite{kharbandacascadexml} have showed that sparse TF–IDF features and dense text features are not mutually exclusive to each other, but rather can be leveraged together as the text representations to boost the performance. Inspired by this strategy, we generate the final text representations by taking advantage of both sparse TF–IDF features and dense vector features, and we propose a novel method to improve the quality of dense vector features for XMC. Specifically, in the fine-tuning stage of pre-trained encoder Transformer, we formulate the multi-label text classification as a text-label matching problem in a bipartite graph. Through text-label alignment and label-text alignment in a bipartite graph, the fine-tuned Transformer can yield robust and effective dense text representations. Besides the dense text representations fine-tuned from the above-mentioned method, we also utilize the static dense sentence embeddings extracted from pre-trained Sentence Transformers, which are widely used in NLP for the tasks, such as text  classification, clustering, retrieval, and paraphrase detection, etc. Compared with the sparse TF-IDF representations, the static dense sentence embeddings can capture the semantic meaning and facilitate the downstream applications. In particular, we extract the static sentence embeddings from Sentence-T5 ~\cite{ni2022sentence} and integrate them into our MatchXML. We have found that this approach is very effective as shown in our ablation study. 

The remainder of the paper is organized as follows. In Section II, we review the related works from the perspectives of extreme classification, cross-modal learning and contrastive learning. The proposed method MatchXML is presented in Section III, where its main components: label2vec, hierarchical label tree, text-label matching, and linear ranker are introduced. Experimental results on six benchmark datasets are presented in Section IV, with comparisons to other algorithms currently in the literature. Conclusions and future work are discussed in Section V.

%Our contributions can be summarized as follows:  (1) We propose \emph{label2vec} to train the semantic dense label embeddings, which are compact and effective to the downstream XMC task. (2) We formulate the multi-label text classification as a text-label matching problem in a bipartite graph. The resulting algorithm MatchXML is simple yet effective to improve the performance. As for classification accuracy,  MatchXML has achieved new state-of-the-art results on five out of six benchmark datasets. As for computational speed, MatchXML has achieved the fastest speed on all the six datasets. 

\section{Related Works}
\label{related_work}

\textbf{Extreme classification. } 
% \subsection{Extreme classification}
A great number of works have been proposed to address the extreme classification problem~\cite{evron2018efficient,
jain2019slice,
jalan2019accelerating,
chalkidis2019large,
medini2019extreme,
prabhu2018extreme,
mittal2021decaf,
dahiya2021deepxml,
mittal2021eclare,
saini2021galaxc,
dahiya2021siamesexml,
gupta2021generalized,
mittal2022multi,
dahiya2023ngame,
babbar2019data,
wydmuch2021propensity,
qaraei2021convex,
schultheis2022speeding,
schultheis2022missing,
jiang2022relevance,
baharav2021enabling,
liu2021label,zong2022bgnn}, which can be categorized to One-vs-All approaches, tree-based approaches, embedding-based approaches, and deep learning approaches. The One-vs-All approaches, such as PDSparse~\cite{yen2016pd}, train a binary classifier for each label independently. To speed up the computation, these approaches leverage the negative sampling and  parallel computing to distribute the training over multiple cores or servers. The tree-based approaches, such as FastXML~\cite{prabhu2014fastxml}, train a hierarchical tree structure to divide the label set  into small groups. These approaches usually have the advantage of fast training and inference.  The embedding-based approaches, such as SLEEC~\cite{bhatia2015sparse},  seek to lower  the computational cost by projecting the high-dimensional label space into a low-dimensional one. However, information loss during the compression process often undermines the classification accuracy. 

Deep learning approaches leverage the raw text to learn semantic dense text representations instead of the statistical TF-IDF features. Recent works (e.g., X-Transformer~\cite{chang2020taming}, APLC-XLNet~\cite{ye2020pretrained}, LightXML~\cite{jiang2021lightxml})  fine-tune the pre-trained encoder Transformers, such as BERT~\cite{kenton2019BERT}, RoBERTa~\cite{liu2019roBERTa} and XLNet~\cite{yang2019xlnet}, to extract the dense text features. Further, a clustering structure or a shallow hierarchical tree structure is designed to deal with the large output label space rather than the traditional linear classifier layer. For example, XR-Transformer~\cite{zhang2021fast} proposes a shallow balanced label tree to fine-tune the pre-trained encoder Transformer in multiple stages. The dense vectors extracted from the last fine-tuning stage and sparse TF-IDF features are leveraged to train the final classifier. Compared with XR-Transformer, we  generate the Hierarchical Label Tree by the label embeddings learned from \emph{label2vec} rather than the TF-IDF features. Besides, we formulate the XMC task as a text-label matching problem to fine-tune the dense text representations. In addition, we extract the static dense sentence embeddings from a pre-trained Sentence Transformer for the classification task. 

\textbf{Cross-Modal Learning. } 
% \subsection{Text matching}
 In the setting of text-label matching, we consider the input texts (i.e., sentences)  as the {\bf text modality}, while the class labels (i.e., 1, 2, 3)  as another {\bf label modality}.   Therefore, the line of research in cross-modal learning is relevant to our text-label matching problem. The cross-modal learning involves processing data across different modalities, such as text, image, audio, and video. Some typical Image-Text Matching tasks have been well studied in recent years, including Image-Text Retrieval~\cite{wang2016learning, lee2018stacked}, Visual Question Answering~\cite{chen2020uniter, tan2019lxmert} and Text-to-Image Generation~\cite{xu2018attngan, zhu2019dm, yin2019semantics, zhang2021cross,  ye2021improving}.  The general framework is to design one image encoder and one text encoder to extract the visual representations and textual representations, respectively, and then fuse the cross-modal information to capture the relationships between them. In contrast to the framework of Image-Text Matching, we develop one text encoder for the text data and one embedding layer to extract the dense label representations. Furthermore, the relationship between image and text in Image-Text Matching usually belongs to an one-to-one mapping, while the relationship between text and label in the context of XMC is a many-to-many mapping. 

\textbf{Contrastive Learning. } 
% \subsection{Contrastive learning}
Another line of research in contrastive learning is also related to our proposed method. Recently, self-supervised contrastive learning~\cite{he2020momentum, chen2020improved, chen2020simple} has attracted great attention due to its remarkable performance in visual representation learning. Typically, a positive pair of images is constructed from two views of the same image, while a negative pair of images is formed from the views of different images. Then a contrastive loss is designed to push together the representations of positive pairs and push apart the ones of negative pairs. Following the framework of self-supervised contrastive learning, supervised contrastive learning~\cite{khosla2020supervised} constructs \emph{additional} positive pairs by utilizing the label information. The application of the supervised contrastive loss can be found in recent works~\cite{chen2022dual, gunel2020supervised, sedghamiz2021supcl, xiong2021extreme} to deal with text classification. In this paper, we leverage the supervised constrastive loss as the training objective for text-label matching, and we develop a novel approach to construct the positive and negative text-label pairs for XMC. MACLR~\cite{ xiong2021extreme} is a recent work that applies the contrastive learning for the Extreme Zero-Shot Learning, and thus is related to our MatchXML. However, there are two main differences between these two works. First, the contrastive learning paradigm in MACLR belongs to self-supervised contrastive learning, while MatchXML is a \emph{supervised} contrastive learning method. Specifically, MACLR constructs the positive text-text pair, where the latter text is a sentence randomly sampled from a long input sentence, while MatchXML constructs the positive text-label pair, where the label is one of the class labels of the input text. Secondly, MACLR utilizes the Inverse Cloze Task which is a frequently used pre-training task for the sentence encoder, while MatchXML is derived from the Cross-Modal learning task.

\section{Method}
\label{Method}

\subsection{Preliminaries}  
%We describe the eXtreme Multi-label text Classification (XMC) in the formal way. 
Given a training dataset with $N$ samples $\{(x_i,y_i)\}_{i=1}^{N}$, where $x_i$ denotes text sample $i$, and $y_i$ is the ground truth that can be expressed as a label vector with binary values of 0 or 1. Let  $y_{il}$, for $l \in \{1,\cdots,L\}$, denote the $l$th element of $ y_i$, where $L$ is the cardinality of the label set. When $y_{il}=1$, label $l$ is relevant to text $i$, and otherwise not. In a typical XMC task, number of instances $N$ and number of labels $L$ can be at the order of millions or even larger. The objective of XMC is to learn a classifier $f(x, l)$ from the training dataset, where the value of $f$ indicates the relevance score of text $x$ and label $l$, with a hope that $f$ can generalize well on test dataset $\{(x_j,y_j)\}_{j=1}^{N_t}$ with a high accuracy.

% The recently proposed XR-Transformer~\cite{zhang2021fast} is an effective method for XMC, which has achieved outstanding performance in terms of accuracy and speed. 
% Following the basic procedure of XR-Transformer~\cite{zhang2021fast}, 
The training of MatchXML consists of four steps. In the first step, we train the dense label vectors by our proposed \emph{label2vec}. In the second step, a preliminary Hierarchical Label Tree (HLT) is constructed using a Balanced K-means Clustering algorithm~\cite{yu2022pecos}. In the third step, a pre-trained Transformer model is fine-tuned recursively from the top layer to bottom layer through the HLT. Finally, we train a linear classifier by utilizing all three text representations: (1) sparse TF-IDF text features, (2) the dense text representations extracted from the fine-tuned Transformer, and (3) the static dense sentence features  extracted from a pre-trained Sentence Transformer. As for the inference, the computational cost contains the feature extraction of input text from the fine-tuned Transformer and the beam search guided by the trained linear classifier through the refined HLT. Thus, the computational complexity of MatchXML inference can be expressed as
\begin{equation}
\label{eqn:inference}
\begin{adjustbox}{max width=195pt}
$
O(T_{1} + k_{b}d\log(L)),
$
\end{adjustbox}
\end{equation}
where $T_{1}$ denotes the cost of extracting the dense text representation from the text encoder, $k_{b}$ is the size of beam search, $d$ is the dimension of the concatenated text representation, and $L$ is the number of labels. The details of MatchXML are elaborated as follows.

\subsection{\emph{label2vec}}
\label{label2vec}
The Hierarchical Label Tree (HLT) plays a fundamental role in  reducing the computational cost of XMC, while the high-quality label embeddings is critical to construct an HLT that can cluster the semantically similar labels together. In this section, we introduce \emph{label2vec} to train the semantic dense label embeddings for the HLT. Note that the training label set $\{y_i\}_{i=1}^{N}$ contains a large amount of semantic information among labels. We therefore treat the positive labels in $y_i$ as a label sequence\footnote{The label order doesn't matter in \emph{label2vec}. Therefore, the label sequence here is actually a label set. However, for easy understanding of \emph{label2vec}, we adopt the same terminology of \emph{word2vec} and treat the positive labels of $y_i$ as a label sequence.}, similar to the words/tokens in one sentence in word2vec. We then adopt the Skip-gram model to train the label embeddings, which can effectively learn high-quality semantic word embeddings from large text corpora. The basic mechanism of the Skip-gram model is to predict context words from a target word. The training objective is to minimize the following loss function:
\begin{equation}
\label{eqn:label2vec}
\begin{adjustbox}{max width=195pt}
$
-\log \sigma(w_t^Tw_c)  -  \sum_{i=1}^{k}  E_{z_i \sim Z_T}\left[\log  \sigma(-w_t^Tw_{z_i}) \right],
$
 \end{adjustbox}
\end{equation}
where $w_t$ and $w_c$ denote the target word embedding and context word embedding, respectively, and $z_i$ is one of the $k$ negative samples. To have the Skip-gram model adapt to the \emph{label2vec} task, we simply make several necessary modifications as follows. First, in \emph{word2vec} the $2n_k$ training target-context word pairs can be generated by setting a context window of size $n_k$, consisting of $n_k$ context words before and after the target word. A small window size (i.e., $n_k$=2) tends to have the target word focusing more on the nearby context words, while a large window size (i.e., $n_k$=10) can capture the semantic relationship between target word and broad context words. The Skip-gram model adopts the strategy of dynamic window size to train \emph{word2vec}. However, in \emph{label2vec} there is no distance constraint between target label and its context labels since they are semantically similar if both labels co-occur in one training sample. Therefore, we set the window size $n_k$ to the maximum number of labels among all training samples. Secondly, the subsampling technique is leveraged to mitigate the imbalance issue between the frequent and rare words in \emph{word2vec} since the frequent/stop words (e.g., “in”, “the”, and “a”) do not provide much semantic information to train word representations. In contrast, the frequent labels are usually as important as rare labels in XMC to capture the semantic relationships among labels in \emph{label2vec}. Therefore, we do not apply the subsampling to the frequent labels in \emph{label2vec}.

\subsection{Hierarchical Label Tree}
\label{hierarchical_labe_tree}
Once the dense label vectors $W =\{w_i\}_{i=1}^{L}$ are extracted from $\{y_i\}^N_{i=1}$ with \emph{label2vec}, we build a Hierarchical Label Tree (HLT) of depth $D$ from the label vectors $W$ by a Balanced K-means Clustering algorithm~\cite{yu2022pecos}. In the construction of HLT, the link relationships of nodes between two adjacent layers are organized as 2D matrices  $C =\{C^t\}_{t=1}^{D}$ based on the clustering assignments. Then the ground truth label assignment of $t$-th layer $Y^{(t)}$ can be generated by the $(t+1)$-th layer $Y^{(t+1)}$ as follows:
\begin{equation}
\label{eqn:layer_label}
 Y^{(t)}=binarize ( Y^{(t+1)} C ^{(t+1)} ).
\end{equation}
The original ground truth $y_i$ corresponds to  $Y^{(D)}$ in the bottom layer, and thus the ground truth label assignment $Y^{(t)}$ can be inferred from the bottom layer to the top layer according to Eq.~\ref{eqn:layer_label}. 
Subsequently, we fine-tune the pre-trained Transformer in multiple stages from the top layer to the bottom layer through the HLT.

\begin{figure*}
\begin{center}
\includegraphics[width=0.80\textwidth]{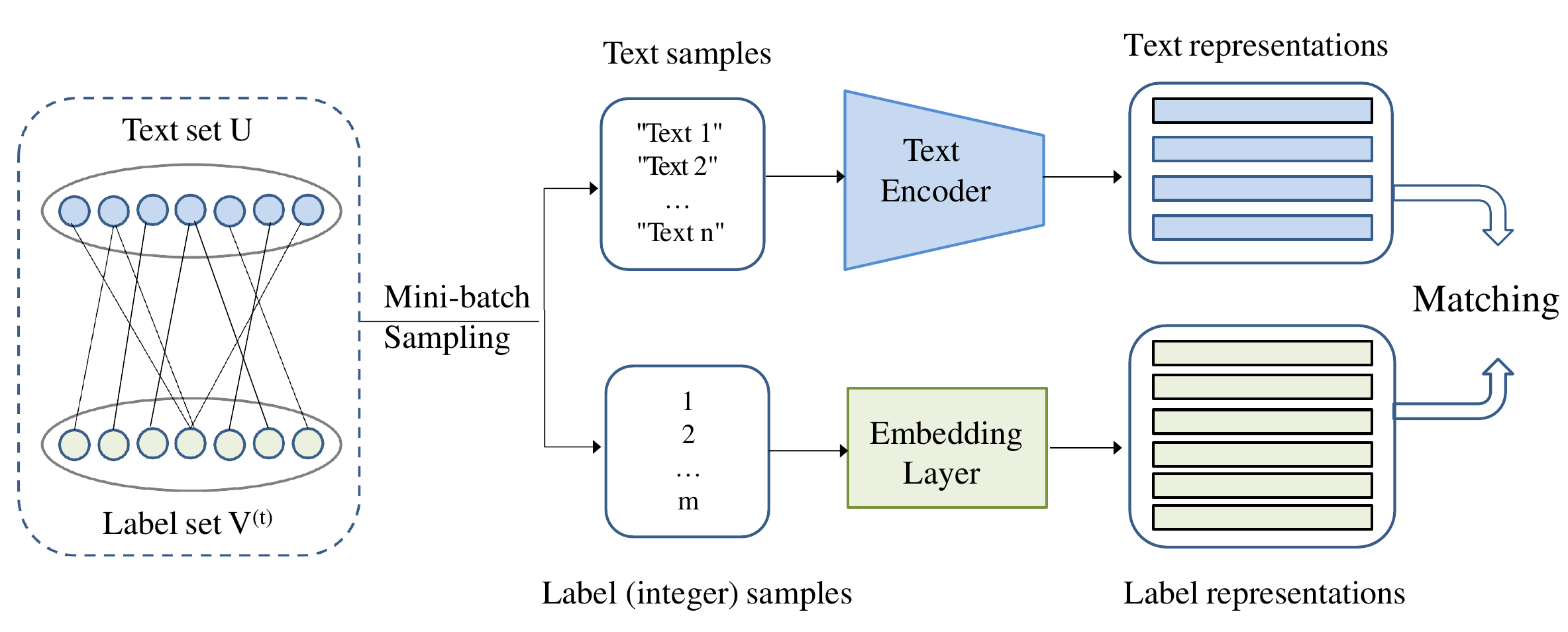}
\end{center}
\vskip -0.2 in
\caption{Architecture of text-label matching in a bipartite graph. When fine-tuning a pre-trained encoder Transformer for the t-th layer of HLT, we consider the input text set U (i.e., training samples)  as the text modality, while the label set $V^{(t)}$ (i.e., training labels $Y^{(t)}$) as the label modality.}
\label{fig:architecture}
 \vskip -0.10in
\end{figure*}

\subsection{Text-label Matching}
\label{text_label}
In this section, we present our text-label matching framework for XMC. In the fine-tuning stage, we consider the multi-label classification as a text-label matching problem. We model this matching problem in a bipartite graph $G(U,V^{(t)},E)$, where $U$ and $V^{(t)}$ denote a set of text samples and the labels in the $t$-th layer of HLT, respectively, and $E$ is a set of edges connecting $U$ and $V^{(t)}$. If text $i$ has a positive label $j$, edge $e_{ij}$ is created between them. A text node in $U$ can have multiple edges connecting it to multiple label nodes in $V^{(t)}$. \emph{Vice versa}, a label node in $V^{(t)}$ can have multiple edges connecting it to multiple text nodes in $U$. We fine-tune a pre-trained encoder Transformer and the HLT from the top layer to the bottom layer in multiple stages. Fig.~\ref{fig:architecture} illustrates the framework of our approach for fine-tuning the encoder Transformer and one layer of the HLT. During training, we sample a mini-batch of training data, from which the text samples are fed to a text encoder to extract the text representations, and the corresponding labels are fed to an embedding layer to extract the label representations. We consider the text-label matching problem from two aspects: text-label alignment and label-text alignment.

\textbf{Text-label alignment. } 
In the text-label matching setting, one text sample aligns with multiple positive labels and contrasts with negative labels in a mini-batch. We construct the set with multiple positive text-label pairs $\{(z_i,e_p)\}$, where $p$ is a positive label of text $i$. Following the previous work~\cite{zhang2021fast}, we also mine the hard negative labels (e.g., negative labels with high output scores) to boost the performance. We then generate the set with a number of negative text-label pairs $\{(z_i,e_n)\}$, where $n$ is one of hard negative labels of text $i$. We utilize the dot product $(z_i,e_j)$ as the quantitative metric to measure the alignment of the text-label pair. To align the text with labels, we train our model to maximize the alignment scores of  positive text-label pairs and minimize the ones of negative text-label pairs. The loss function of text-label alignment is defined as
% \begin{equation}
% \label{eqn:cond-gan}
% \begin{adjustbox}{max width=215pt}
% $
% \underset{G}{\min}\:
% \underset{D}{\max} \!\! \underset{\:\:x \sim P_r}{\mathbb{E}}\![\log(D(x, e))]\!+\!\!\!\underset{\:z \sim P_z}{\mathbb{E}} [\log(1\!-\!D(G(z,e),e))] 
%  $
%  \end{adjustbox}
% \end{equation}
\begin{equation}
\label{eqn:text_label}
\begin{adjustbox}{max width=215pt}
$
 \mathcal{L}_{tl}=  \frac{1}{N_b} \sum_{i=1}^{N_b}
 \frac{1}{|P_1(i)|} \sum_{p\in P_1(i)}
 -\log\frac{\exp((z_i, e_p)/\tau)}  {  \displaystyle \sum_{a \in A_1(i) } \exp((z_i, e_a)/\tau)  }, 
$ 
 \end{adjustbox}
\end{equation}
where ${N_b}$ denotes the batch size,  $P_1 (i)$ is the set of indices of positive labels related to text $i$, $|P_1(i)|$ is its cardinality, $A_1(i)$ is the set of indices of positive and negative labels corresponding to text $i$, and $\tau \in R^+ $ is a
scalar temperature parameter.

\textbf{Label-text alignment. } 
We also consider the label-text alignment in a reverse way for the text-label matching problem. In the above-mentioned text-label alignment, we mine a number of hard negative labels for each text to facilitate the training process. On the contrary, if we form the label set by combining all the positive labels and hard negative labels within a mini-batch, the computational cost is likely to increase notably due to the large cardinality of the label set. To reduce the computational cost, we construct the label set \emph{only} from all the positive labels within a mini-batch. Similar to the previous text-label alignment, one label sample corresponds to several text samples and contrasts with the remaining text samples in the mini-batch. We generate the set with several positive label-text pairs $\{(e_i,z_p)\}$, where $i$ is a positive label for text $p$. Otherwise, they form the set with a number of negative label-text pairs $\{(e_i,z_n)\}$, where $i$ is a negative label for text $n$. To align the label with texts, we train our model to maximize the alignment scores of positive label-text pairs and minimize the ones of negative label-text pairs. Similarly, the loss function of label-text alignment is defined as
\begin{equation}
\label{eqn:label_text}
\begin{adjustbox}{max width=215pt}
$ 
\mathcal{L}_{lt}=  \frac{1}{M} \sum_{i=1}^{M}
 \frac{1}{|P_2 (i)|} \sum_{p\in P_2 (i)}
 -\log\frac{\exp((e_i, z_p)/\tau)}  {  \displaystyle \sum_{a \in A_2 (i) } \exp((e_i, z_a)/\tau)  },
 $ 
 \end{adjustbox}
\end{equation}
where $M$ is the number of positive labels in the mini-batch, $P_2(i)$ is the set of indices of positive text samples related to label $i$, $|P_2(i)|$ is its cardinality, and $A_2(i)$ is the set of indices of text samples within the mini-batch.

\textbf{Loss function.} The overall loss function of our text-label matching task is a linear combination of the two loss functions defined above
\begin{equation}
\label{eqn:loss}
\mathcal{L} = \lambda \mathcal{L}_{tl} +  (1- \lambda ) \mathcal{L}_{lt},
\end{equation}
with $\lambda$ $\in [0,1] $. Experiments show that the setting of hyperparameter $\lambda$ has a notable impact on the performance of MatchXML, and we thus tune it for different datasets.

\begin{table*}[t]
\begin{center}

\caption{Statistics of datasets. $N_{train}$ is the number of training samples, $N_{test}$ is the number of test samples, $D$ is the dimension of  feature vector, $L$ is the cardinality of  label set, $\bar{L}$ is the average number of labels per sample, $\hat{L}$ is
the average samples per label. }
\vspace{-0.50em}

\begin{adjustbox}{width=0.75\textwidth}
% \vskip -0.05in
\label{table:dataset}
\begin{tabular}{c r r r r r r }
\hline
Dataset & $N_{train}$ & $N_{test}$ & $D$ & $L$ & $\bar{L}$ & $\hat{L}$ \\
\hline
EURLex-4k & 15,449 & 3,865 & 186,104 & \textbf{3,956} & 5.30 & 20.79   \\
AmazonCat-13k & 1,186,239 & 306,782 & 203,882 & \textbf{13,330} & 5.04 & 448.57   \\
Wiki10-31k & 14,146 & 6,616 & 101,938 & \textbf{30,938} & 18.64 & 8.52  \\
Wiki-500k & 1,779,881 & 769,421 & 2,381,304 & \textbf{501,070} & 4.75 & 16.86   \\
Amazon-670k & 490,449 & 153,025 & 135,909 & \textbf{670,091} & 5.45 & 3.99  \\
Amazon-3M & 1,717,899 & 742,507 & 337,067 & \textbf{2,812,281} & 36.04 & 22.02   \\
\hline
\end{tabular}

% \vskip -0.1 in

\end{adjustbox}
\end{center}

% \vskip -0.25in
\end{table*}

\begin{figure*}
\begin{center}
\includegraphics[width=0.95\textwidth]{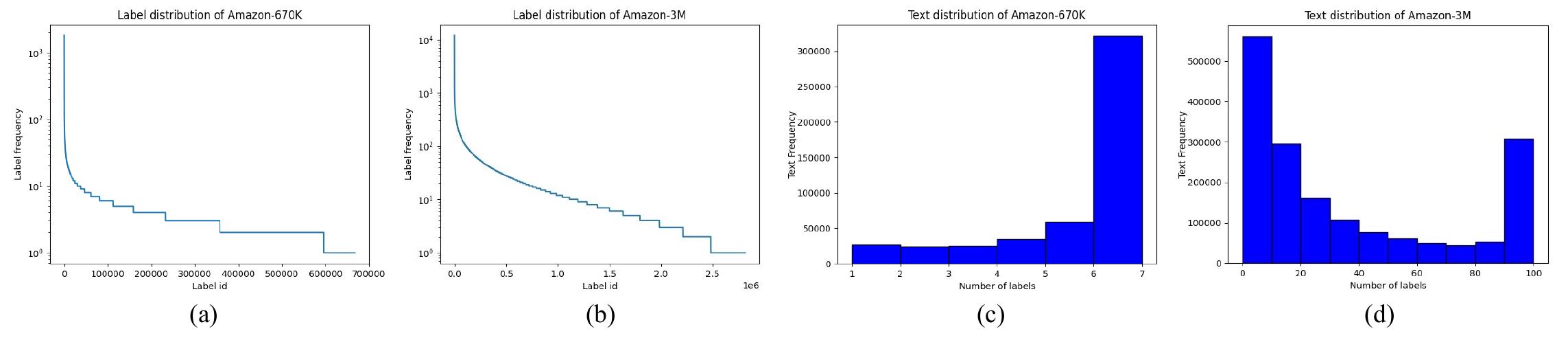}
\end{center}
\vskip -0.2 in
\caption{Label distributions of Amazon-670K and Amazon-3M follow the power (Zipf’s) Law, as shown in (a) and (b). Text distributions of Amazon-670K and Amazon-3M don't follow a particular standard form, as shown in (c) and (d).}
\label{fig:distribution}
 \vskip -0.10in
\end{figure*}

\begin{figure}[t]
\vspace{-0.5em}
\begin{algorithm}[H]  
\small
\caption{MatchXML Training} \label{alg:training}  
\begin{algorithmic}[1]
\REQUIRE  {Training dataset $ \{ X, Y \}=\{(x_i,y_i)\}_{i=1}^{N}$, TF-IDF features $ \{ \bar{X} \}=\{(\bar{x}_i)\}_{i=1}^{N}$,  static dense sentence embeddings $ \{ \check{X} \}=\{(\check{x}_i)\}_{i=1}^{N}$, Skip-gram model $h$,  text encoder $g$, the depth of HLT $D$}
\ENSURE{Optimized text encoder $g$ and the hierarchical linear ranker  $\{{R}^{(t)}\}_{t=1}^{{D}}$ }

\STATE 
Generate label pairs $\{(l_i^k, l_j^k)\}_{k=1}^{K}$ from $ \{ Y \}=\{y_i\}_{i=1}^{N}$ 

\STATE 
{\bf for} \{1, $\cdots$, \#  of  training epochs\} do \\

\STATE 
\ \ \  {\bf for}  \{1, $\cdots$, \#  of  training steps\} do \\

\STATE 
\ \ \ \ \ \ Sample a mini-batch of label pairs  $\{(l_i, l_j)\}$

\STATE 
\ \ \ \ \ \ Update Skip-gram model $h$ to minimize Eq.~\ref{eqn:label2vec}
\STATE 
\ \ \ {\bf end for}\\

\STATE 
{\bf end for}\\

\STATE 
Obtain dense label vectors $  W =\{w_i\}_{i=1}^{L}$  $\leftarrow$ $ h$  \\

\STATE 
  $\{{C}^{(t)}\}_{t=1}^{{D}}$ $\leftarrow$ Balanced K-means Clustering($ W $)  \\
  
 \STATE 
  Get  hierarchical ground truth label assignment $\{{Y}^{(t)}\}_{t=1}^{{D}}$ by Eq.~\ref{eqn:layer_label} \\

\STATE 
{\bf for} \{1, $\cdots$, $D$ \} do \\

\STATE 
\ \ \ Initialize label embedding layer  $E^{(t)}$ by the Bootstrap

\STATE 
\ \ \  {\bf for}  \{1, $\cdots$, \#  of  training steps\} do \\

\STATE 
\ \ \ \ \ \ Sample a mini-batch of training  samples  $\{ (x_i, y_i^{(t)} \}$  \\
\STATE 
\ \ \ \ \ \ Construct text-label pairs  $\{ (z_i, e_j \}$ \\

\STATE 
\ \ \ \ \ \   Construct label-text pairs  $\{ (e_{\hat{i}},z_{\hat{j}})\}$ \\

\STATE 
\ \ \ \ \ \ Update  Encoder $g$ and  Embedding  $E^{(t)}$ to  minimize Eq.~\ref{eqn:loss}  \\

\STATE 
\ \ \ {\bf end for}\\

\STATE 
{\bf end for}\\

\STATE 
Obtain dense text features $ \hat {X} =\{\hat{x}_i\}_{i=1}^{N}$  $\leftarrow$ $ g(X) $  \\

\STATE 
Obtain final text features $ \tilde { X } =\{\tilde {x}_i\}_{i=1}^{N}$  $\leftarrow$ $ Concat(\hat {X}, \bar{X}, \check{X} ) $  \\

\STATE 
{\bf for} \{1, $\cdots$, $D$ \} do \\

\STATE 
\ \ \ Train the linear ranker ${R}^{(t)}$ of the t-th layer by Eq. ~\ref{eqn:ranker}

\STATE 
{\bf end for}\\

\end{algorithmic}

\end{algorithm}

\vspace{-1.5em}
\end{figure}

 \subsection{Linear Ranker}
\label{text_label}
Once the multi-stage fine-tuning with Eq. (\ref{eqn:loss}) is completed, we extract the dense text representations from the text encoder. The extracted dense representations are then concatenated with the static dense sentence embeddings from the Sentence Transformer and the sparse TF-IDF features as the final text representations $\{\tilde{x}_i\}_{i=1}^{N}$, which are used to train a linear ranking model based on XR-LINEAR~\cite{yu2022pecos}. Specifically, let  $W^{(t)}$ denote the learnable parameter matrix of the ranker corresponding to the $t$-th layer of HLT, $\hat{M}^{(t)}$ denote the matrix of sampled labels by the combination of the Teacher-Forcing Negatives (TFN) and Matcher-Aware Negatives (MAN), $Y^{(t)}$ denote the label assignment at the $t$-th layer of HLT. The linear ranker at the $t$-th layer can be optimized as:
\begin{equation}\label{eqn:ranker}
%\begin{adjustbox}{max width=195pt}
{
 \arg\min_{W^{(t)}} \sum_{\ell: \hat{M}^{(t)}_{i,\ell}\neq 0} 
 \mathcal{L} ( Y_{i \ell }^{(t)}, W ^{(t)\top}_\ell 
  \tilde{x}_i)) + \alpha \|W^{(t)} \|^2,
}
%\end{adjustbox}
\end{equation}
where $\alpha$ is the hyperparameter that balances the classification loss and the $L_2$ regularization on the parameter matrix $W^{(t)}$.

In summary, the training procedure of MatchXML is provided in Algorithm~\ref{alg:training}.

\begin{table*}
% \vskip -0.05in
\caption{Settings of hyperparameters to train  \emph{label2vec} on six datasets. $w_{size}$ denotes the window size, $ns$ denotes exponent value used to shape the negative sampling distribution, $n_{epoch}$ is the number of training epochs, $dim$ is the dimension of label vector, $n_{neg}$ is the number of negative labels, $lr_{max}$ and $lr_{min}$ denote the maximum and minimum learning rate in the training process, respectively, $r_{sample}$ is the downsampling threshold, and $sg=1$ refers to Skip-gram model.}

\vspace{-0.90em}

\begin{center}
% \vskip +0.1in
\begin{adjustbox}{width=0.80\textwidth}
\begin{tabular}{ l | c c c |  c | c | c | c | c | c }
\toprule
%  {} & \multicolumn{3}{|c|}{Stage I}  & \multicolumn{3}{|c|}{Stage II}& \multicolumn{3}{|c| }{Stage III}   \\
% \hline
Dataset &  $w_{size}$ & $ns$ &  $n_{epoch} $ &  $dim$ & $n_{neg}$ & $lr_{max}$  &  $lr_{min}$ &   $r_{sample}$ & $sg$   \\
\midrule

Eurlex-4K & 24 &  0.5 &  20 &   \multirow{6}{*}  {100} &   \multirow{6}{*}  {20} &   \multirow{6}{*}  {2.5e-2}  &  \multirow{6}{*}  {1e-4}  &  \multirow{6}{*}  {0.1}  &  \multirow{6}{*}  {$1$}   \\

Wiki10-31K & 30 &  1.0 &  20 &   &    &   &   &   &    \\
AmazonCat-13K & 57 &  0.5 &  20 &   &   &   &   &   &     \\
Wiki-500K & 274 &  -1.0 &  50 &   &   &   &   &   &     \\
Amazon-670K & 7 &  0.5 &  50 &   &   &   &   &   &     \\

Amazon-3M & 100 &  -0.5 &  20 &   &   &   &   &   &    \\

\bottomrule
\end{tabular}
\end{adjustbox}

\end{center}
% \vskip -0.05in
\label{table:vector_param}
% \vskip -0.2in
\end{table*}

\begin{table*}[t]
% \vskip -0.25in

% \vskip -0.05in

\caption{Setting of learning rates and training steps  on six datasets. $lr_t $ and $lr_l $ denotes the learning rate of the text encoder and embedding layer, respectively. $n_{step}$ denotes the number of training steps. }

\vspace{-0.90em}

\begin{center}
% \vskip +0.1in
\begin{adjustbox}{width=0.90\textwidth}
\begin{tabular}{ l | c c c|  c c c| c  c c | c c c }
\toprule
 {} & \multicolumn{3}{|c|}{Stage \upperRomannumeral{1}}  & \multicolumn{3}{|c|}{Stage \upperRomannumeral{2}}& \multicolumn{3}{|c| }{Stage \upperRomannumeral{3}} &
 \multicolumn{3}{|c}{Stage \upperRomannumeral{4} }
 \\
\hline
Dataset &  $lr_t$ & $lr_l$ & $n_{step}$  & $lr_t$  &  $lr_l$ &   $n_{step}$ & $lr_t$  & $lr_l$ & $n_{step}$  &  $lr_t$  & $lr_l$ & $n_{step}$\\
\midrule

Eurlex-4K & 5e-5 &  1e-3 &  480 &  5e-5 &  1e-3 &  620 &  5e-5 &  1e-3 &  600  & -- & -- &  -- \\

Wiki10-31K & 5e-5 &  1e-3 &  500 &  5e-5 &  1e-3  &  520 &  5e-5 &  1e-3 &  350 & -- & -- &  --\\
AmazonCat-13K & 1e-4 &  1e-3 &  10,000 &  1e-4 &  1e-3 &  10,000 &  1e-4 &  1e-3 &  20,000  & -- & -- &  --\\
Wiki-500K & 1e-4 &  1e-3 &  10,000 &  1e-4 &  1e-3 &  10,000 &  1e-4 &  1e-3 &  20,000  & 1e-4 & 1e-3 &  20,000\\
Amazon-670K & 5e-5 &  1e-3 &  4,000 &  5e-5 &  1e-3 &  4,000 &  2e-4 &  1e-3 &  12,000  & -- & -- &  --\\

Amazon-3M & 1.5e-4 &  5e-3 &  10,000 &  1.5e-4 &  5e-3 &  10,000 &  1.5e-4 &  5e-3 &  10,000  & -- & -- &  --\\

\bottomrule
\end{tabular}
\end{adjustbox}

\end{center}

% \vskip -0.1in
\vspace{-0.75em}

\label{table:lr}
% \vskip -0.2in
\end{table*}

\begin{table*}[h]

% \vskip -0.05in

\caption{Setting of hyperparameters for fine-tuning on six datasets. $NB_{trn}$ and $NB_{tst}$ denote the batch size for training and inference, respectively.  $Length$ refers to the sequence length. $\tau$ is a
scalar temperature defined in Eq.~\ref{eqn:text_label} and Eq.~\ref{eqn:label_text}. $\lambda$ is the coefficient defined in Eq.~\ref{eqn:loss}. $emb$ denotes which type of label embeddings is leveraged for the construction of Hierarchical Label Tree.  $w_d$,  $betas$,   $eps$ denotes the value of weight decay, betas and epsilon, respectively, for the optimizer. }

\vspace{-0.90em}

\begin{center}
% \vskip +0.1in
\begin{adjustbox}{width=0.90\textwidth}
\begin{tabular}{ l | c | c |c |c |  c |c |c |c |c |c}
\toprule
%  {} & \multicolumn{3}{|c|}{Stage \upperRomannumeral{1}}  & \multicolumn{3}{|c|}{Stage \upperRomannumeral{2}}& \multicolumn{3}{|c| }{Stage \upperRomannumeral{3}} &
%  \multicolumn{3}{|c}{Stage \upperRomannumeral{4} }
%  \\
% \hline
Dataset &  $Encoder$ & $NB_{trn}$ &    $NB_{tst}$ & $Length$ & $\tau$  & $\lambda$  &  $emb$  &  $w_d$ &  $betas$ &  $eps$\\
\midrule

Eurlex-4K & BERT &  128 &  256 &  128 &  0.05 &  1.0 & TF-IDF &  0.1 & (0.9,0.98)  & 1e-6 \\

Wiki10-31K & RoBERTa &  128 &  256 &  256 &  0.05  &  0 & \emph{label2vec} &  0.1 & (0.9,0.98)  & 1e-6 \\
AmazonCat-13K & BERT &  128 &  256 &  256 &  0.05 &  1.0 & \emph{label2vec} &  0.1 & (0.9,0.98)  & 1e-6\\
Wiki-500K & BERT &  128 &  256 &  128 &  0.05 &  0.67 & \emph{label2vec}  &  0.1 & (0.9,0.98)  & 1e-6\\
Amazon-670K & RoBERTa &  256 &  512 & 128 &  0.05 &  0.5 & \emph{label2vec} &  0.1 & (0.9,0.98)  & 1e-6 \\

Amazon-3M & BERT &  256 &  512 &  128 & 0.05 &  0.9 & \emph{label2vec} &  0.1 & (0.9,0.98)  & 1e-6\\

\bottomrule
\end{tabular}
\end{adjustbox}

\end{center}

\vspace{-0.50em}

% \vskip -0.1in
\label{table:param}
% \vskip -0.2in
\end{table*}

\begin{table*}[t]
% \begin{table}[t]
% \vskip -0.25in

\caption{Comparison of our approach with recent XMC methods on six public datasets.The best result among all the methods is in bold. The symbol $P@k$ denotes the evaluation metric defined in Eq.~\ref{eqn:pk}.   The symbol ``\textunderscore "    denotes the second best result. The symbol ``$*$ " refers to our reproduced result.  The symbol ``--"  denotes that the result is not provided in the original paper.}

\vspace{-0.75em}
 % \vskip -0.8in

\begin{center}
% \vskip +0.1in
\begin{adjustbox}{width=0.95\textwidth}
\begin{tabular}{ l | l l  l | l  l  l | l l  l }
\toprule
 {} & \multicolumn{3}{|c|}{Eurlex-4K}  & \multicolumn{3}{|c|}{Wiki10-31K}  &  \multicolumn{3}{c}{AmazonCat-13K} \\
\hline
Method & P@1 & P@3  & P@5  &  P@1 & P@3  & P@5 & P@1 & P@3  & P@5\\
\midrule
AnnexML~\cite{tagami2017annexml} & 79.66 & 69.64 & 53.52 & 86.46  & 74.28 & 64.20 & 93.54 & 78.36 & 63.30 \\
DiSMEC~\cite{babbar2017dismec} & 83.21 & 70.39 & 58.73 & 84.13 & 74.72 & 65.94 & 93.81 & 79.08 & 64.06 \\
PfastreXML~\cite{jain2016extreme} & 73.14  & 60.16 &  50.54 &  83.57 &  68.61 &  59.10  & 91.75  & 77.97 &  63.68 \\
Parabel~\cite{prabhu2018parabel} & 82.12 &  68.91 &  57.89  & 84.19 &  72.46 &  63.37 &  93.02 &  79.14 &  64.51 \\
eXtremeText~\cite{wydmuch2018no} & 79.17  & 66.80 &  56.09 &  83.66 &  73.28 &  64.51 &  92.50 &  78.12 &  63.51 \\
Bonsai~\cite{khandagale2019bonsai} & 82.30 &  69.55 &  58.35 &  84.52 &  73.76 &  64.69 &  92.98 &  79.13 &  64.46 \\

XR-Linear~\cite{yu2022pecos} & 84.14 &  72.05 &  60.67 &  85.75 &  75.79 &  66.69 &  94.64 &  79.98 &  64.79 \\
\hline
XML-CNN~\cite{liu2017deep} & 75.32 &  60.14 &  49.21 &  81.41 &  66.23 &  56.11 &  93.26 &  77.06 &  61.40 \\
AttentionXML~\cite{you2019attentionxml} & 85.49 &  73.08 &  61.10 &  87.10 &  77.80 &  68.80 &  95.65 &  81.93 &  66.90 \\

LightXML~\cite{jiang2021lightxml} & 86.02$^*$ &  74.02$^*$ &  \underline{61.87}$^*$ &  87.80 &  77.30 &  68.00 &  \textbf{96.55}$^*$ &  \textbf{83.70}$^*$ &  \textbf{68.46}$^*$ \\
APLC-XLNet~\cite{ye2020pretrained} & 83.60$^*$ & 70.20$^*$ & 57.90$^*$ & \underline{88.76}$^*$  & \underline{79.11}$^*$ & \underline{69.63}$^*$ & 96.14$^*$ & 82.86$^*$ &67.58$^*$ \\
XR-Transformer~\cite{zhang2021fast} & \underline{87.22}$^*$ &  \underline{74.39}$^*$ &  61.69$^*$ &  88.00 &  78.70 &  69.10 &  96.25$^*$ &  82.72$^*$ &  67.01$^*$ \\
MatchXML(ours) & \textbf{88.12}{\footnotesize(+0.90)} & \textbf{75.00}{\footnotesize(+0.61)} & \textbf{62.22}{\footnotesize(+0.35)} & \textbf{89.30}{\footnotesize(+0.54)}  & \textbf{80.45}{\footnotesize(+1.34)} & \textbf{70.89}{\footnotesize(+1.26)} & \underline{96.50}{\footnotesize(-0.05)} & \underline{83.25}{\footnotesize(-0.45)} &\underline{67.69}{\footnotesize(-0.77)} \\
\hline
\hline
 {} & \multicolumn{3}{|c|}{Wiki-500K}  & \multicolumn{3}{|c|}{Amazon-670K}  &  \multicolumn{3}{c}{Amazon-3M} \\
\hline
Method & P@1 & P@3  & P@5  &  P@1 & P@3  & P@5 & P@1 & P@3  & P@5\\
\midrule
AnnexML~\cite{tagami2017annexml} & 64.22 &  43.15 &  32.79 &  42.09 &  36.61 &  32.75 &  49.30 &  45.55 &  43.11 \\
DiSMEC~\cite{babbar2017dismec} & 70.21  & 50.57  & 39.68  & 44.78 &  39.72 &  36.17 &  47.34 &  44.96 &  42.80 \\
PfastreXML~\cite{jain2016extreme} & 56.25 &  37.32 &  28.16 &  36.84 &  34.23 &  32.09 &  43.83 &  41.81 &  40.09 \\
Parabel~\cite{prabhu2018parabel} & 68.70  & 49.57  & 38.64 &  44.91  & 39.77 &  35.98 &  47.42 &  44.66 &  42.55 \\
eXtremeText~\cite{wydmuch2018no} & 65.17 &  46.32 &  36.15 &  42.54 &  37.93 &  34.63 &  42.20 &  39.28 &  37.24 \\
Bonsai~\cite{khandagale2019bonsai} & 69.26 &  49.80 &  38.83 &  45.58 &  40.39 &  36.60 &  48.45 &  45.65 &  43.49 \\

XR-Linear~\cite{yu2022pecos} & 65.59 &  46.72 &  36.46 &  43.38 &  38.40 &  34.77 &  47.40 &  44.15 &  41.87 \\
\hline
XML-CNN~\cite{liu2017deep} & -- &  -- &  -- &  33.41 &  30.00 &  27.42 &  -- &  -- &  -- \\
AttentionXML~\cite{you2019attentionxml} & 75.10 &  56.50 &  44.40 &  45.70 &  40.70 &  36.90 &  49.08 &  46.04 &  43.88 \\

LightXML~\cite{jiang2021lightxml} & 76.30 & 57.30 & 44.20 &  47.30 &  42.20 &  38.50&  		--  &  -- &  -- \\
APLC-XLNet~\cite{ye2020pretrained} & 75.47$^*$ & 56.84$^*$ & 44.20$^*$ & 43.54$^*$  & 38.91$^*$ & 35.33$^*$ & -- & -- & -- \\
XR-Transformer~\cite{zhang2021fast} & \underline{78.10} &  \underline{57.60} &  \underline{45.00} &  \underline{49.10} & \underline{43.80} & \underline{40.00} &  \underline{52.60} &   \underline{49.40} &   \underline{46.90} \\
MatchXML (ours) & \textbf{79.80}{\footnotesize(+1.70)} & \textbf{59.28}{\footnotesize(+1.68)} & \textbf{46.03}{\footnotesize(+1.03)} & \textbf{50.83}{\footnotesize(+1.73)}  & \textbf{45.37}{\footnotesize(+1.57)} & \textbf{41.30}{\footnotesize(+1.30)} & \textbf{54.22}{\footnotesize(+1.62)} & \textbf{50.84}{\footnotesize(+1.44)} & \textbf{48.27}{\footnotesize(+1.37)} \\

\bottomrule
\end{tabular}

\end{adjustbox}
\end{center}

% \vskip -0.02in

% \vspace{-0.50em}

% \vskip 0.05in
\label{table:p1_result}
% \vskip -0.2in
\end{table*}

\section{Experiments}
%\subsection{Datasets}

We conduct experiments to evaluate the performance of MatchXML on six public datasets~\cite{Bhatia16}\footnote{\url{https://ia802308.us.archive.org/21/items/pecos-dataset/xmc-base/}}, including EURLex-4K, Wiki10-31K, AmazonCat-13K, Wiki-500K, Amazon-670K, and Amazon-3M, which are the same datasets used by XR-Transformer~\cite{zhang2021fast}. The statistics of these datasets can be found in Table~\ref{table:dataset}. It is well-known that the label distribution of the XMC datasets follows the power (Zipf’s) law, where most of the probability mass is covered by a small fraction of the label set. As for the text distribution, each document is categorized by a different number of labels, and this distribution doesn’t follow a particular standard form. This can be observed from Fig.~\ref{fig:distribution}, where the label and text distributions of Amazon-670K and Amazon-3M are provided.

We consider EURLex-4K, Wiki10-31K, and AmazonCat-13K as medium-scale datasets, while Wiki-500K, Amazon-670K, and Amazon-3M as large-scale datasets. We are more interested in the performance on large-scale datasets since they are more challenging XMC tasks.

\vspace{-10pt}
\subsection{Evaluation Metrics}
%In this section, we define the evaluation metrics used in this paper. 
% The precision @ k @ k and nDCG @ k @ k metrics are defined for a predicted score vector ^ y ∈ R L y ^ ∈ R L and ground truth label vector y ∈ { 0 , 1 } L y ∈ { 0 , 1 } L as
The widely used evaluation metrics for XMC are the precision at $k$ (P@k) and ranking quality at $k$ (nDCG@k), which are defined as
\begin{align}
    \label{eqn:pk}
  \text{P@k}&=\frac{1}{k}\sum_{l \in rank_{k}(\hat{\mathbf{y}})} \mathbf{y}_{l},\\
  \label{eqn:dcg}
  \text{DCG@k}&=\frac{1}{k}\sum_{l \in rank_{k}(\hat{\mathbf{y}})} \frac {\mathbf{y}_{l}}{log(l+1)},\\
    \label{eqn:ndcg}
\text{nDCG@k}&=\frac{\text{DCG@k}}{\sum_{l=1}^{min(k, ||\mathbf{y}_{0}||)} \frac{1}{log(l+1)}},
\end{align}
where $\mathbf{y}\in\{0,1\}^L$ is the ground truth label, $\hat{\mathbf{y}}$ is the predicted score vector, and $rank_{k}(\hat{\mathbf{y}})$ returns the $k$ largest indices of $\hat{\mathbf{y}}$, sorted in descending order.

For datasets that contain a large percentage of head (popular) labels, high P@k or nDCG@k may be achieved by simply predicting well on head labels.  
For performance evaluation on tail (infrequent) labels,  the XCM methods are recommended to evaluate with respect to the propensity-scored counterparts of the precision P@k and nDCG metrics (PSP and PSnDCG), which are defined as
\begin{align}
  \label{eqn:pspk}
  \text{PSP@k}&=\frac{1}{k}\sum_{l \in rank_{k}(\hat{\mathbf{y}})} \frac {\mathbf{y}_{l}}{p_l},\\
  \label{eqn:psdcgk}
  \text{PSDCG@k}&=\frac{1}{k}\sum_{l \in rank_{k}(\hat{\mathbf{y}})} \frac {\mathbf{y}_{l}}{p_l \, log(l+1)},\\
   \label{eqn:psndcgk}
\text{PSnDCG@k}&=\frac{\text{PSDCG@k}}{\sum_{l=1}^{k} \frac{1}{log(l+1)}},
\end{align}
where $p_{l}$ is the propensity score of label $l$ that is used to make metrics unbiased with respect to missing labels~\cite{Bhatia16}.  For
consistency, we use the same setting as XR-Transformer~\cite{zhang2021fast} for all datasets.

Following the prior works, we also record the Wall-clock time of our program for speed comparison.

\begin{table*}[t]
% \vskip -0.05in
\caption{Comparison of our approach and baselines in terms of ensemble model. The results of the baselines are cited from XR-Transformer. The symbol $P@k$ denotes the evaluation metric defined in Eq.~\ref{eqn:pk}. The suffix $-3$ and $-9$ denote the ensemble model has three or nine models, respectively. The underline symbol ``\textunderscore" denotes the second best result.}

\vspace{-0.75em}

\begin{center}
% \vskip +0.1in
\begin{adjustbox}{width=0.95\textwidth}
\begin{tabular}{ l | l l  l | l  l  l | l l  l }
\toprule
 {} & \multicolumn{3}{|c|}{Eurlex-4K}  & \multicolumn{3}{|c|}{Wiki10-31K}  &  \multicolumn{3}{c}{AmazonCat-13K} \\
\hline
Method & P@1 & P@3  & P@5  &  P@1 & P@3  & P@5 & P@1 & P@3  & P@5\\
\midrule

AttentionXML-3 ~\cite{you2019attentionxml} & 86.93 & 74.12 & 62.16 & 87.34 & 78.18 & 69.07 & 95.84 & 82.39 & 67.32 \\

X-Transformer-9~\cite{chang2020taming} & 87.61 & 75.39 & 63.05 & 88.26 & 78.51 & 69.68 & 96.48 & 83.41 & 68.19 \\

LightXML-3~\cite{jiang2021lightxml}  & 87.15 & 75.95 & \textbf{63.45} & \underline{89.67} & 79.06 & 69.87 & 96.77 & \textbf{83.98} & \textbf{68.63} \\

XR-Transformer-3~\cite{zhang2021fast}  & \underline{88.41} & \underline{75.97} & 63.18 & 88.69 & \underline{80.17} & \underline{70.91} & \underline{96.79} & 83.66 & 68.04 \\
MatchXML-3(ours) & \textbf{88.85}{\footnotesize(+0.44)} & \textbf{76.02}{\footnotesize(+0.05)} & \underline{63.30}{\footnotesize(-0.15)} & \textbf{89.74}{\footnotesize(+0.07)}  & \textbf{81.51}{\footnotesize(+1.34)} & \textbf{72.18}{\footnotesize(+1.27)} & \textbf{96.83}{\footnotesize(+0.04)} & \underline{83.83}{\footnotesize(-0.15)} & \underline{68.20}{\footnotesize(-0.43)} \\
\hline
\hline
 {} & \multicolumn{3}{|c|}{Wiki-500K}  & \multicolumn{3}{|c|}{Amazon-670K}  &  \multicolumn{3}{c}{Amazon-3M} \\
\hline
Method & P@1 & P@3  & P@5  &  P@1 & P@3  & P@5 & P@1 & P@3  & P@5\\
\midrule

AttentionXML-3~\cite{you2019attentionxml}  & 76.74 & 58.18 & 45.95 & 47.68 & 42.70 & 38.99 & 50.86 & 48.00 & 45.82 \\
X-Transformer-9~\cite{chang2020taming}  & 77.09 & 57.51 & 45.28 & 48.07 & 42.96 & 39.12 & 51.20 & 47.81 & 45.07 \\

LightXML-3~\cite{jiang2021lightxml} & 77.89 & 58.98 & 45.71 & 49.32 & 44.17 & 40.25 &   -   &   -   &   -   \\

XR-Transformer-3~\cite{zhang2021fast}  & \underline{79.40} & \underline{59.02} & \underline{46.25} & \underline{50.11} & \underline{44.56} & \underline{40.64} & \underline{54.20} & \underline{50.81} & \underline{48.26} \\
MatchXML-3(ours) & \textbf{80.66}{\footnotesize(+1.26)} & \textbf{60.43}{\footnotesize(+1.41)} & \textbf{47.09}{\footnotesize(+0.84)} & \textbf{51.64}{\footnotesize(+1.53)}  & \textbf{46.17}{\footnotesize(+1.61)} & \textbf{42.05}{\footnotesize(+1.41)} & \textbf{55.88}{\footnotesize(+1.68)} & \textbf{52.39}{\footnotesize(+1.58)} & \textbf{49.80}{\footnotesize(+1.54)} \\

\bottomrule
\end{tabular}
\end{adjustbox}

\end{center}
\vskip 0.1in
\label{table:ensemble}
% \vskip -0.2in
% \end{table}
\end{table*}

\begin{table*}[t]
% \begin{table}[t]
% \vskip -0.25in

\caption{Comparison of our approach with recent XMC methods on six public datasets.The best result among all the methods is in bold. The symbol ``\textunderscore" denotes the second best result. The symbol $nDCG@k$ denotes the evaluation metric defined in Eq.~\ref{eqn:ndcg}. 
The symbol ``--" denotes that the result is not provided in the original paper. The suffix $-3$  denotes the ensemble model has three models. }

\vspace{-0.75em}
 % \vskip -0.8in

\begin{center}
% \vskip +0.1in
\begin{adjustbox}{width=0.95\textwidth}
\begin{tabular}{ l | l l  l | l  l  l | l l  l }
\toprule
 {} & \multicolumn{3}{|c|}{Eurlex-4K}  & \multicolumn{3}{|c|}{Wiki10-31K}  &  \multicolumn{3}{c}{AmazonCat-13K} \\
\hline
Method & nDCG@1 & nDCG@3  & nDCG@5  &  nDCG@1 & nDCG@3  & nDCG@5 & nDCG@1 & nDCG@3  & nDCG@5\\
\midrule
AnnexML~\cite{tagami2017annexml} & 79.26  & 68.13 & 61.60 &  86.49 & 77.13 &  69.44 &  93.54  &  87.29 & 85.10 \\
DiSMEC~\cite{babbar2017dismec} & 82.40 & 72.50 &  66.70 & 84.10 &  77.10  &  70.40 &  93.40 &  87.70&  85.80 \\
PfastreXML~\cite{jain2016extreme} & 76.37 & 66.63 &  60.61 &  83.57  &  72.00 &  64.54 &  91.75 &  86.48 & 84.96 \\
Parabel~\cite{prabhu2018parabel} & 82.25 & 72.17 &  66.54 &  84.17 & 75.22 &  68.22 &  93.03  & 87.72 &  86.00 \\
% eXtremeText~\cite{wydmuch2018no} & 0.0 &  0.0 &  0.0 &  0.0 &  0.0 &  0.0 &  0.0 &  0.0 &  0.0 \\
Bonsai~\cite{khandagale2019bonsai} & 82.96 &  73.15 &  67.41 &  84.69 & 76.25 & 69.17 &  92.98 & 87.68 & 85.92 \\

% XR-Linear~\cite{yu2022pecos} & 0.0 &  0.0 &  0.0 &  0.0 &  0.0 &  0.0 &  0.0 &  0.0 &  0.0 \\
% \hline
XML-CNN~\cite{liu2017deep} & 76.38 &  66.28 & 60.32 &  81.42 &  69.78 & 61.83 &  93.26 & 86.20 & 83.43 \\
AttentionXML-3~\cite{you2019attentionxml} & 87.12 & 77.44 & 71.53 & 87.47  & 80.61 & 73.79 &  95.92 & 91.97 & 89.48 \\

% LightXML~\cite{jiang2021lightxml} & 0.0 &  0.0 &  0.0 &  0.0 &  0.0 &  0.0 &  0.0 &  0.0 &  0.0 \\
% APLC-XLNet~\cite{ye2020pretrained}& 0.0 &  0.0 &  0.0 &  0.0 &  0.0 &  0.0 &  0.0 &  0.0 &  0.0 \\
XR-Transformer-3~\cite{zhang2021fast}& \underline{88.20} &  \underline{79.29} &  \underline{72.98} &  \underline{88.75} &  \underline{82.37} &  \underline{75.38} &  \underline{96.71} &  \underline{92.39} &  \underline{90.31} \\
MatchXML-3(ours) & \textbf{88.85}{\footnotesize(+0.65)} & \textbf{79.50}{\footnotesize(+0.21)} & \textbf{73.26}{\footnotesize(+0.28)} & \textbf{89.74}{\footnotesize(+0.99)}  & \textbf{83.46}{\footnotesize(+1.09)} & \textbf{76.53}{\footnotesize(+1.15)} & \textbf{96.83}{\footnotesize(+0.12)} & \textbf{92.59}{\footnotesize(+0.20)} & \textbf{90.62}{\footnotesize(+0.31)} \\
\hline
\hline
 {} & \multicolumn{3}{|c|}{Wiki-500K}  & \multicolumn{3}{|c|}{Amazon-670K}  &  \multicolumn{3}{c}{Amazon-3M} \\
\hline
Method & nDCG@1 & nDCG@3  & nDCG@5  &  nDCG@1 & nDCG@3  & nDCG@5 & nDCG@1 & nDCG@3  & nDCG@5\\
\midrule
AnnexML~\cite{tagami2017annexml} & 64.22 & 54.30 & 52.23 &  42.39 &  39.07 &  37.04 &  49.30 &  46.79 &  45.27 \\
DiSMEC~\cite{babbar2017dismec} & 70.20 &  42.10 &  40.50 &  44.70 &  42.10 &  40.50 & \textendash &  \textendash &  \textendash \\
PfastreXML~\cite{jain2016extreme} & 59.20 &  30.10 &  28.70 &  39.46 &  37.78 &  36.69 &  43.83 &  42.68 &  41.75 \\
Parabel~\cite{prabhu2018parabel} & 67.50 & 38.50 &  36.30 &  44.89 &  42.14 &  40.36 &  47.48 &  45.73 &  44.53 \\
% eXtremeText~\cite{wydmuch2018no} & 0.0 &  0.0 &  0.0 &  0.0 &  0.0 &  0.0 &  0.0 &  0.0 &  0.0 \\
Bonsai~\cite{khandagale2019bonsai} & 69.20 &  60.99 &  59.16 &  45.58 &  42.79 &  41.05 &  48.45 &  46.78 &  45.59 \\

% XR-Linear~\cite{yu2022pecos} & 0.0 &  0.0 &  0.0 &  0.0 &  0.0 &  0.0 &  0.0 &  0.0 &  0.0 \\
% \hline
XML-CNN~\cite{liu2017deep} & 69.85 & 58.46 & 56.12 &  35.39 &  33.74 &  32.64 &  \textendash &  \textendash &  \textendash \\
AttentionXML-3~\cite{you2019attentionxml} & 76.95 &  70.04  & 68.23 &  47.58 &  45.07 &  43.50 &  50.86 &  49.16 &  47.94 \\

% LightXML~\cite{jiang2021lightxml} & 0.0 &  0.0 &  0.0 &  \textendash &  \textendash &  \textendash &  \textendash &  \textendash &  \textendash \\
% APLC-XLNet~\cite{ye2020pretrained}& 0.0 &  0.0 &  0.0 &  0.0 &  0.0 &  0.0 &  \textendash &  \textendash &  \textendash \\
XR-Transformer-3~\cite{zhang2021fast} & \underline{79.43} &  \underline{71.74} &  \underline{69.88} &  \underline{50.01} & \underline{47.20} & \underline{45.51} &  \underline{54.22} &   \underline{52.29} &   \underline{50.97} \\
MatchXML-3 (ours) & \textbf{80.66}{\footnotesize(+1.23)} & \textbf{73.28}{\footnotesize(+1.54)} & \textbf{71.20}{\footnotesize(+1.32)} & \textbf{51.63}{\footnotesize(+1.62)}  & \textbf{48.81}{\footnotesize(+1.61)} & \textbf{47.04}{\footnotesize(+1.53)} & \textbf{55.88}{\footnotesize(+1.66)} & \textbf{53.90}{\footnotesize(+1.61)} & \textbf{52.58}{\footnotesize(+1.61)} \\

\bottomrule
\end{tabular}

\end{adjustbox}
\end{center}

% \vskip -0.02in

% \vspace{-0.50em}

% \vskip 0.05in
\label{table:ensemble_ndcg}
% \vskip -0.2in
\end{table*}

\subsection{Experimental Settings}
\label{sec_setting}

We train the dense label embeddings by using the Skip-gram model of the Gensim library, which contains an efficient implementation of \emph{word2vec} as described in the original paper~\cite{mikolov2013distributed}. We take the label sequences $\{y_i\}_{i=1}^{N}$ of training data as the input corpora, and set the dimension of label vector to $100$ and number of negative label samples to $20$. In \emph{word2vec}, some rare words would be ignored if the frequency is less than a certain threshold. We keep all the labels in the label vocabulary regardless of the frequency. The settings of the Skip-gram model for the six datasets are listed in Table~\ref{table:vector_param}.

Following the prior works, we utilize BERT~\cite{devlin2018BERT} as the major text encoder in our experiments. Instead of using the same learning rate for the whole model, we leverage the discriminative learning rate~\cite{howard2018universal, ye2020pretrained} to fine-tune our model, which assigns different learning rates for the text encoder and the label embedding layer. Following XR-Transformer, we use different optimizers AdamW~\cite{loshchilov2017decoupled} and SparseAdam for the text encoder and the label embedding layer, respectively. Since the size of parameters in the label embedding layer can be extremely large for large datasets, the SparseAdam optimizer is utilized to reduce the GPU memory consumption and improve the training speed. Further, prior Transformer-based approaches~\cite{chang2020taming,jiang2021lightxml,ye2020pretrained} have shown that the longer input text usually improves classification accuracy, but leads to more expensive computation. However, we find that the classification accuracy of MatchXML is less sensitive to the length of input text since MatchXML utilizes both dense feature vectors extracted from Transformer and the TF-IDF features for classification. We therefore truncate the input text to a reasonable length to balance the accuracy and speed. In the multi-stage fine-tuning process, we only apply the proposed text-label matching learning in the last stage, while we keep the original multi-label classification learning for the other fine-tuning stages. 
As shown in Table~\ref{table:lr}, we set different learning rates for the text encoder and the label embedding layer in each fine-tuning stage. There is a three-stage process for fine-tuning the Transformer on five datasets, including Eurlex-4K, Wiki10-31K, AmazonCat-13K, Amazon-670K, and Amazon-3M, and a four-stage process on Wiki-500K. Table~\ref{table:param} provides the further details of the hyperparameters. We extract the static sentence embeddings from the pre-trained Sentence-T5 model~\cite{ni2022sentence}.

\begin{table*}[t]
% \begin{table}[t]
    \centering
    \caption{Comparison of  our approach and baselines on three large datasets w.r.t. $PSP@k$.  The symbol $PSP@k$ denotes the evaluation metric defined in Eq.~\ref{eqn:pspk}.The symbol ``$*$" refers to our reproduced result. The underline symbol ``\textunderscore" denotes the second best result. The suffix $-3$  denotes the ensemble model has three models. The results of XR-Transformer and MatchXML are from singe model.}
    
    \vspace{-0.25em}
    
    %\resizebox{0.90\textwidth}{!}{
    \resizebox{0.95\textwidth}{!}{
    \begin{tabular}{llllllllll}
        \toprule
        & \multicolumn{3}{c}{ Wiki-500K } & \multicolumn{3}{c}{ Amazon-670K }     & \multicolumn{3}{c}{ Amazon-3M } \\
        \hline
        \hline
        Method & PSP@1 & PSP@3 & PSP@5 & PSP@1 & PSP@3 & PSP@5 & PSP@1 & PSP@3 & PSP@5\\
        \midrule
        Pfastrexml~\cite{jain2016extreme}      & 32.02 & 29.75 & 30.19 & 20.30 & 30.80 & 32.43 & \textbf{21.38} & \textbf{23.22} & \textbf{24.52} \\
        Parabel~\cite{prabhu2018parabel}       & 26.88 & 31.96 & 35.26 & 26.36 & 29.95 & 33.17 & 12.80 & 15.50 & 17.55 \\
        \midrule
        XR-Transformer-3~\cite{zhang2021fast}                 & \underline{35.45}$^*$ & \underline{42.39}$^*$ & \underline{46.74}$^*$ & \underline{29.88}$^*$ & \underline{34.31}$^*$ & \underline{38.54}$^*$ & 16.61$^*$ & 20.06$^*$ & 22.55$^*$ \\
        MatchXML-3(ours)                & \textbf{35.87}{\footnotesize(+0.42)} & \textbf{43.12}{\footnotesize(+0.73)} & \textbf{47.50}{\footnotesize(+0.76)} & \textbf{30.30}{\footnotesize(+0.42)} & \textbf{35.28}{\footnotesize(+0.97)} & \textbf{39.78}{\footnotesize(+1.24)} & \underline{17.00}{\footnotesize(-4.38)} & \underline{20.55}{\footnotesize(-2.67)} & \underline{23.16}{\footnotesize(-1.36)} \\
        \bottomrule
    \end{tabular}
    }
	\label{table:psp-result}
% \end{table}

\vspace{0.25em}

\end{table*}

\begin{table*}[t]

\caption{Training time (in \textbf {hours}) of single model of our approach and recent deep learning methods on six public datasets. The symbol ``$*$" and ``--" have the same meanings as in Table~\ref{table:p1_result}. }

\vspace{-0.75em}

\begin{center}
% \vskip +0.1in
\begin{adjustbox}{width=0.90\textwidth}
% {width=0.95\textwidth}
\begin{tabular}{ l | c c  c  c  c  c  c }
\toprule
Method {} & Eurlex-4K  & Wiki10-31K  & AmazonCat-13K & Wiki-500K  & Amazon-670K  & Amazon-3M \\
\hline
%  & P1 & P3  & P5  &  P1 & P3  & P5 & P1  \\
% \midrule
AttentionXML~\cite{you2019attentionxml} & 0.30 & 0.50 & 8.1 & 12.5  & 8.1  & 18.27  \\
X-Transformer~\cite{chang2020taming} & 0.83 & 1.57 & 16.4 &  61.9 & 57.2  & 60.2  \\
LightXML~\cite{jiang2021lightxml} & 5.63 & 8.96 & 103.6 & 90.4  & 53.0  & --  \\
APLC-XLNet~\cite{ye2020pretrained} & 3.15$^*$ & 2.16$^*$ & 43.2$^*$ & 118.9$^*$  & 63.0$^*$  & --  \\

XR-Transformer~\cite{zhang2021fast} & 0.26 & 0.50 & 13.2 & 12.5 & 3.4  & 9.7   \\
MatchXML (ours) & \textbf{0.20} & \textbf{0.22} & \textbf{6.6} & \textbf{11.1}  & \textbf{3.3}  & \textbf{8.3}  \\

\bottomrule
\end{tabular}
\end{adjustbox}

\end{center}
\label{table:matchxml_time}
\vspace{-5pt}
\end{table*}

\begin{table*}[t]
\caption{Training time (in \textbf {hours}) of \emph{label2vec}.}
\vspace{-10pt}
\begin{center}
\begin{adjustbox}{width=0.90\textwidth}
\begin{tabular}{ l | c c  c  c  c  c  c }
\toprule
 {} & Eurlex-4K  & Wiki10-31K  & AmazonCat-13K & Wiki-500K  & Amazon-670K  & Amazon-3M \\
\hline
%  & P1 & P3  & P5  &  P1 & P3  & P5 & P1  \\
% \midrule
\emph{label2vec} & 0.01 & 0.02 & 0.05 & 0.27  & 0.08  & 3.60  \\
\bottomrule
\end{tabular}
\end{adjustbox}
\end{center}
\label{table:label2vec_time}
\vspace{-5pt}
\end{table*}

\begin{table*}[t]
\caption{Comparison of the embedding sizes (in \textbf {Mb}) between TF-IDF features and dense vectors on six datasets.}
\vspace{-10pt}
\begin{center}
\begin{adjustbox}{width=0.90\textwidth}
\begin{tabular}{ l | c| c | c | c | c|  c }
\toprule
 {} & Eurlex-4K  & Wiki10-31K  & AmazonCat-13K & Wiki-500K  & Amazon-670K  & Amazon-3M \\
\hline
%  & P1 & P3  & P5  &  P1 & P3  & P5 & P1  \\
% \midrule
TF-IDF & 29.1 & 344.5 & 179.9 & 7,109.2  & 783.8  & 7,422.4  \\
\emph{label2vec} & 1.6 & 12.4 & 5.1 & 200.4  & 268.0  & 1,073.0  \\
\bottomrule
\end{tabular}
\end{adjustbox}
\end{center}
\label{table:label2vec_size}
\vspace{-5pt}
\end{table*}

We compare our MatchXML with 12 state-of-the-art (SOTA) XMC methods:
AnnexML~\cite{tagami2017annexml}, DiSMEC~\cite{babbar2017dismec}, PfastreXML~\cite{jain2016extreme}, Parabel~\cite{prabhu2018parabel}, eXtremeText~\cite{wydmuch2018no}, Bonsai~\cite{khandagale2019bonsai}, XML-CNN~\cite{liu2017deep}, XR-Linear~\cite{yu2022pecos}, AttentionXML~\cite{you2019attentionxml}, LightXML~\cite{jiang2021lightxml}, APLC-XLNet~\cite{ye2020pretrained}, and XR-Transformer~\cite{zhang2021fast}. For deep learning approaches (XML-CNN, AttentionXML, LightXML, APLC-XLNet, XR-Transformer, MatchXML), we list the results of the single model for a fair comparison. We also provide the results of ensemble model.
The results of the baseline methods are cited from the XR-Transformer paper. For parts of the results that are not available in XR-Transformer, we reproduce the results using the source code provided by the original papers. The original paper of APLC-XLNet has reported the results of another version of datasets, which are different from the ones in XR-Transformer. We therefore reproduce the results of APLC-XLNet by running the source code on the same datasets as XR-Transformer. Our experiments were conducted on a GPU server with 8 Tesla V100 GPUs and 64 CPUs, which has the same number of Tesla V100 GPUs and CPUs as the AWS p3.16xlarge utilized by XR-Transformer.

\subsection{Experimental Results}

\textbf{Classification accuracy. }
Table~\ref{table:p1_result} shows the classification accuracies of our MatchXML and the baseline methods over the six datasets. Overall, MatchXML has achieved state-of-the-art results on five out of six datasets. Especially, on three large-scale datasets: Wiki-500K, Amazon-670K, and Amazon-3M, and the gains are about $1.70 \%$,  $1.73\%$ and $1.62\%$ in terms of P@1, respectively, over the second best results. Compared with the baseline XR-Transformer, MatchXML has a better performance in terms of precision on all the six datasets. For AmazonCat-13K, our approach has achieved the second best result, with the performance gap of 0.05\% compared with LightXML. Note that the number of labels for this dataset is not large (about 13K), indicating that it can be handled reasonably well by the linear classifier in LightXML, while our hierarchical structure is superior when dealing with datasets with extremely large label outputs.

\textbf{Results of ensemble models. }
% \section{Results of ensemble models}
\label{sec_ensemble}
We have the similar ensemble strategy as XR-Transformer. That is, three pre-trained text encoders (BERT, RoBERTa, XLNet) are utilized together as the ensemble model for three small datasets, including Eurlex-4K, Wiki10-31K, and AmazonCat-13K; and one text encoder with three different Hierarchical Label Trees are formed the ensemble model for three large datasets, including Wiki-500K, Amazon-670K, and Amazon-3M. As shown in Table~\ref{table:ensemble}, our MatchXML again achieves state-of-the-art results on four datasets: Wiki10-31K, Wiki-500K, Amazon-670K, and Amazon-3M in terms of three metrics P@1, P@3 and P@5, which is consistent with the results of the single model setting. For dataset Eurlex-4K, P@1 and P@3 of our approach are the best, similar to the single model results, while the P@5 is the second best with a slight performance gap of 0.15, compared with the best result. For AmazonCat-13K, P@1 of our approach achieves the best result, while P@3 and P@5 are the second best, similar to the ones in single model.

Table~\ref{table:ensemble_ndcg} shows the performances in terms of the ranking metric nDCG@k of our MatchXML and the baselines over the six datasets. Similarly, our MatchXML achieves state-of-the-art results on all the six datasets. For the three medium-scale datasets, Eurlex-4K, Wiki10-31K, AmazonCat-13K, the performance gains of nDCG@1 are about $0.65\%$, $1.0\%$ and $0.12\%$ over the second best results, respectively. For the three large-scale datasets: Wiki-500K, Amazon-670K and Amazon-3M, the gains are about $1.23\%$, $1.62\%$ and $1.66\%$ over the second best results, respectively.

% \begin{table*}[h]
% % \vskip -0.05in
% \caption{Comparison of P@k between TF-IDF and \emph{label2vec} on six datasets.}

% \vspace{-0.75em}

% \begin{center}
% % \vskip +0.1in
% \begin{adjustbox}{width=0.90\textwidth}
% \begin{tabular}{ l | c c  c | c  c  c | c c  c }
% \toprule
%  {} & \multicolumn{3}{|c|}{Eurlex-4K}  & \multicolumn{3}{|c|}{Wiki10-31K}  &  \multicolumn{3}{c}{AmazonCat-13K} \\
% \hline
% Method & P@1 & P@3  & P@5  &  P@1 & P@3  & P@5 & P@1 & P@3  & P@5\\
% \midrule

% MatchXML (w/ TF-IDF)  & \textbf{88.18} &  75.05 &  62.38 &  89.13 &  80.18 &  70.48 &  \textbf{96.48} &  \textbf{83.26} &  \textbf{67.69} \\
% MatchXML (w/ \emph{label2vec})  & 87.79 & \textbf{75.36} & \textbf{62.56} & \textbf{89.24}  & \textbf{80.49} & \textbf{70.57} & \textbf{96.48} & 83.24 & \textbf{67.69} \\
% \hline
% \hline
%  {} & \multicolumn{3}{|c|}{Wiki-500K}  & \multicolumn{3}{|c|}{Amazon-670K}  &  \multicolumn{3}{c}{Amazon-3M} \\
% \hline
% Method & P@1 & P@3  & P@5  &  P@1 & P@3  & P@5 & P@1 & P@3  & P@5\\
% \midrule

% MatchXML (w/ TF-IDF)  & 78.89 &  58.12 &  45.28 &  50.04 & 44.61 & 40.72 &  53.05 &   49.74 &   47.24 \\
% MatchXML (w/ \emph{label2vec})  & \textbf{79.14} & \textbf{58.56} & \textbf{45.51} & \textbf{50.40}  & \textbf{44.99} & \textbf{41.00} & \textbf{53.27} & \textbf{49.96} & \textbf{47.44} \\

% \bottomrule
% \end{tabular}
% \end{adjustbox}

% \end{center}
% % \vskip -0.1in

% \vspace{0.75em}

% % \vskip -0.05in
% \label{table:label2vec_p1}
% % \vskip -0.2in
% \end{table*}

\textbf{Results of propensity scored precision. }
% \section{Results of propensity scored precision}
\label{sec_psp}
We compute the propensity scored precision (PSP@k) to measure the performance of MatchXML on tail labels. The results of the baselines: PfastreXML and Parabel are cited from the official website\footnote{\url{http://manikvarma.org/downloads/XC/XMLRepository.html}}. The results reported in the XR-Transformer paper are computed using a different version of source code. We reproduce the results of XR-Transformer and compute the  PSP@k using the official source code\footnote{\url{https://github.com/kunaldahiya/pyxclib}}. As shown in Table~\ref{table:psp-result}, our MatchXML again achieves state-of-the-art results on two out of three large datasets Wiki-500K and Amazon-670K in terms of three metrics PSP@1, PSP@3 and PSP@5. For Amazon-3M, our approach has achieved the second best performance. Note that Parabel has developed  specific techniques to boost the performance on tail labels, and thus has the best performance on tail labels of Amazon-3M. However, as shown in Table~\ref{table:p1_result}, the performance of Parabel on all the labels is about 6\% lower than our approach.

\begin{table*}[t]
% \vskip -0.05in

\caption{Ablation study of our MatchXML. ``l2v" refers to the label2vec method,``tlm" refers to the text label matching method and ``sen" denotes the static sentence embeddings. The symbol $P@k$ denotes the evaluation metric defined in Eq.~\ref{eqn:pk}. The best result among all the methods is in bold. 
% The blue color denotes the performance increase, while the orange color denotes the performance drop. 
}

\vspace{-0.75em}

\begin{center}
% \vskip +0.1in
\begin{adjustbox}{width=0.90\textwidth}
\begin{tabular}{ | l | l | l  | l  | l l  l | }
\toprule
 \multirow{2}{*}{} & \multirow{2}{*}{l2v} & \multirow{2}{*}{tlm} &
\multirow{2}{*}  {sen} &
 \multicolumn{3}{|c|}{Eurlex-4K}    \\
\cline{5-7}
 & & & &  P@1 & P@3  & P@5   \\
\midrule

1 &  &   &  & \textbf{87.99} &  74.76 &  61.98  \\
2& $\surd$  & &  & 87.35  \textcolor{black}{(- 0.64)} & 74.89 \textcolor{black}{(+ 0.13)}   & 62.05 \textcolor{black}{(+ 0.07)}    \\

3& $\surd$ &  $\surd$ &   & 87.66 \textcolor{black}{(\qquad + 0.31)}  & \textbf{75.27}  \textcolor{black}{(\qquad + 0.38)}   & \textbf{62.54} \textcolor{black}{(\qquad + 0.49)}    \\

4 \qquad  & $\surd$ \qquad \quad   & $\surd$ \qquad  \quad & $\surd$ \qquad  \quad & 87.87 \textcolor{black}{(\qquad \qquad   + 0.21)}    & 74.94  \textcolor{black}{(\qquad \qquad   - 0.33)}  & 62.25 \textcolor{black}{(\qquad \qquad   - 0.29)}    \\

\hline
\hline
 \multirow{2}{*}{} & \multirow{2}{*}{l2v} & \multirow{2}{*}{tlm} &
\multirow{2}{*}  {sen}  & \multicolumn{3}{|c|}{Wiki10-31K}    \\
\cline{5-7}
 & & & &  P@1 & P@3  & P@5   \\
\midrule

1&  &   &  & 88.80 &  80.17 &  70.41  \\
2& $\surd$ & &  & 89.10  \textcolor{black}{(+ 0.30)} & 80.22 \textcolor{black}{(+ 0.05)}   & 70.69 \textcolor{black}{(+ 0.28)}    \\

3& $\surd$ &  $\surd$ &   & 89.21 \textcolor{black}{(\qquad + 0.11)}  & 80.13  \textcolor{black}{(\qquad - 0.09)}   & 70.24 \textcolor{black}{(\qquad - 0.45)}    \\

4& $\surd$ & $\surd$ & $\surd$  & \textbf{89.30} \textcolor{black}{(\qquad \qquad   + 0.09)}    & \textbf{80.45}  \textcolor{black}{(\qquad \qquad   + 0.32)}  & \textbf{70.89} \textcolor{black}{(\qquad \qquad   + 0.65)}    \\

\hline
\hline
 \multirow{2}{*}{} & \multirow{2}{*}{l2v} & \multirow{2}{*}{tlm} &
\multirow{2}{*}  {sen}  & \multicolumn{3}{|c|}{AmazonCat-13K}   \\
\cline{5-7}
 & & & &  P@1 & P@3  & P@5  \\
\midrule

1&  &   &  & 96.42 &  83.18 &  67.63  \\
2& $\surd$ & &  & 96.41  \textcolor{black}{(- 0.01)} & 83.19 \textcolor{black}{(+ 0.01)}   & 67.63 \textcolor{black}{(+ 0)}    \\

3& $\surd$ &  $\surd$ &   & 96.48 \textcolor{black}{(\qquad + 0.07)}  & 83.24  \textcolor{black}{(\qquad + 0.05)}   & 67.69 \textcolor{black}{(\qquad + 0.06)}    \\

4& $\surd$ & $\surd$ & $\surd$  & \textbf{96.50} \textcolor{black}{(\qquad \qquad   + 0.02)}    & \textbf{83.25}  \textcolor{black}{(\qquad \qquad   + 0.01)}  & \textbf{67.69} \textcolor{black}{(\qquad \qquad   + 0)}    \\

\hline
\hline
 \multirow{2}{*}{} & \multirow{2}{*}{l2v} & \multirow{2}{*}{tlm} &
\multirow{2}{*}  {sen}  & \multicolumn{3}{|c|}{Wiki-500K}   \\
\cline{5-7}
 & & & &  P@1 & P@3  & P@5  \\
\midrule

1&  &   &  & 78.65 &  58.02 &  45.24  \\
2& $\surd$ & &  & 78.84  \textcolor{black}{(+ 0.19)} & 58.46 \textcolor{black}{(+ 0.44)}   & 45.49 \textcolor{black}{(+ 0.25)}    \\

3& $\surd$ &  $\surd$ &   & 79.14 \textcolor{black}{(\qquad + 0.30)}  & 58.56  \textcolor{black}{(\qquad + 0.10)}   & 45.51 \textcolor{black}{(\qquad + 0.02)}    \\

4& $\surd$ & $\surd$ & $\surd$  & \textbf{79.80} \textcolor{black}{(\qquad \qquad   + 0.66)}    & \textbf{59.28}  \textcolor{black}{(\qquad \qquad   + 0.72)}  & \textbf{46.03} \textcolor{black}{(\qquad \qquad   + 0.52)}    \\

\hline
\hline
 \multirow{2}{*}{} & \multirow{2}{*}{l2v} & \multirow{2}{*}{tlm} &
\multirow{2}{*}  {sen}  & \multicolumn{3}{|c|}{Amazon-670K}   \\
\cline{5-7}
 & & & &  P@1 & P@3  & P@5  \\
\midrule

1&  &   &  & 49.33 &  43.91 &  40.01  \\
2& $\surd$ & &  & 49.53  \textcolor{black}{(+ 0.20)} & 44.21 \textcolor{black}{(+ 0.30)}   & 40.36 \textcolor{black}{(+ 0.35)}    \\

3& $\surd$ &  $\surd$ &   & 50.44 \textcolor{black}{(\qquad + 0.91)}  & 44.94  \textcolor{black}{(\qquad + 0.73)}   & 41.00 \textcolor{black}{(\qquad + 0.64)}    \\

4& $\surd$ & $\surd$ & $\surd$  & \textbf{50.83} \textcolor{black}{(\qquad \qquad   + 0.39)}    & \textbf{45.37}  \textcolor{black}{(\qquad \qquad   + 0.43)}  & \textbf{41.30} \textcolor{black}{(\qquad \qquad   + 0.30)}    \\

\hline
\hline
 \multirow{2}{*}{} & \multirow{2}{*}{l2v} & \multirow{2}{*}{tlm} &
\multirow{2}{*}  {sen}  & \multicolumn{3}{|c|}{Amazon-3M}   \\
\cline{5-7}
 & & & &  P@1 & P@3  & P@5  \\
\midrule

1&  &   &  & 52.92 &  49.67 &  47.22  \\
2& $\surd$ & &  & 53.11  \textcolor{black}{(+ 0.19)} & 49.88 \textcolor{black}{(+ 0.21)}   & 47.39 \textcolor{black}{(+ 0.17)}    \\

3& $\surd$ &  $\surd$ &   & 53.27 \textcolor{black}{(\qquad + 0.16)}  & 49.96  \textcolor{black}{(\qquad + 0.08)}   & 47.44 \textcolor{black}{(\qquad + 0.05)}    \\

4& $\surd$ & $\surd$ & $\surd$  & \textbf{54.22} \textcolor{black}{(\qquad \qquad   + 0.95)}    & \textbf{50.84}  \textcolor{black}{(\qquad \qquad   + 0.88)}  & \textbf{48.27} \textcolor{black}{(\qquad \qquad   + 0.83)}    \\

\bottomrule
\end{tabular}
\end{adjustbox}

\end{center}
\vspace{-0.75em}

 % \vskip -0.075in
\label{table:ablation_alignment}
% \vskip -0.2in
\end{table*}

\textbf{Computation Cost.}
Table~\ref{table:matchxml_time} reports the training costs of our MatchXML and other deep learning based approaches. The baseline results of training time are cited from XR-Transformer. For the unavailable training time of a single model, we calculate it by dividing the training time of ensemble model by the number of models in the ensemble. In XR-Transformer, $13.2$ hour is the reported training time of the ensemble of three models for AmazonCat-13K. We have checked the sequence length (which is 256) and the number of training steps (which is 45,000). We believe this cost should be the training time of single model. Overall, our approach has shown the fastest training speed on all the six datasets. We fine-tune the text encoder in three stages from the top layer to the bottom layer through the HLT. Furthermore, we leverage several training techniques, such as discriminative learning rate, small batch size and less training steps, to improve the convergence rate of our approach. Our MatchXML has the same strategy for inference as XR-Transformer. The inference time on six datasets can be found in Appendix  A.4.2 of XR-Transformer~\cite{zhang2021fast}. 

\subsection{Ablation study}
The framework of our MatchXML follows the training procedure as the baseline XR-Transformer, including the construction of HLT, fine-tuning the encoder Transformer from the top to bottom layers through the HLT, and training a linear classifier. Besides, we have proposed three novel techniques to boost the performance, namely \emph{label2vec} to learn the dense label embeddings for the HLT construction, text-label matching for fine-tuning the encoder Transformer, and extraction of static dense text embeddings from pre-trained Sentence Transformer. In the ablation study, we set up our technical contribution one by one and report the experimental results to show the effectiveness of each component. The performance of our base model is comparable to or slightly better than the baseline XR-Transformer, since we have leveraged some techniques to speed up the training.

\textbf{Performance of label2vec. } Table~\ref{table:ablation_alignment} reports the performance comparison of \emph{label2vec} (number $2$) and TF-IDF (number $1$) in terms of precision for the downstream XMC tasks. On the small datasets, e.g., Eurlex-4K, Wiki10-31K and AmazonCat-13K, the performances of label embeddings from \emph{label2vec} are comparable to the ones from TF-IDF features. However, on the large datasets, e.g., Wiki-500K, Amazon-670K and Amazon-3M, \emph{label2vec} outperforms TF-IDF, indicating that a large training corpus is essential to learn high-quality dense label embeddings. Our experimental results show that \emph{label2vec} is more effective than TF-IDF to utilize the large-scale datasets.

Table~\ref{table:label2vec_time} reports the training time of \emph{label2vec} on the six datasets. The training of \emph{label2vec} is highly efficient on five of them, including Eurlex-4K, Wiki10-31K, AmazonCat-13K, Wiki-500K and Amazon-670K, as the cost is less than $0.3$ hours. The training time on Amazon-3M is about $3.6$ hours, which is the result of large amount of training label pairs. As shown in Table~\ref{table:dataset}, the number of instances $N_{train}$ and the average number of positive labels per instance $\overline L$ are the two factors that determine the size of training corpus. Note that we do not add the training time of \emph{label2vec} into the classification task since we consider the \emph{label2vec} task as the preprocessing step for the downstream tasks.

Table~\ref{table:label2vec_size} compares the sizes of label embedding from \emph{label2vec} and TF-IDF. The dense label vectors have much smaller size than that of the sparse TF-IDF label representations. Especially, on the large dataset, such as Wiki-500K, the size of label embeddings can be reduced by $35\times$ (from $7,109.2$MB to $200.4$MB), which benefits the construction of HLT significantly.

\textbf{Performance of text-label matching. }
Table~\ref{table:ablation_alignment} reports the performance of our text-label matching (number $3$) on the six datasets. The baseline objective is the weighted squared hinge loss~\cite{zhang2021fast} (number $2$). Our text-label matching approach outperforms the baseline method on five out of six datasets, including Eurlex-4K, Amazoncat-13K, Wiki-500K, Amazon-670K and Amazon-3M. For Wiki10-31K, the metric P@1 is still better than the baseline, while P@3 and P@5 are slightly worse. On the three large-scale datasets, the text-label matching has achieved the largest gain of about 0.91\% on Amazon-670K, while the small gain of about 0.16\% on Amazon-3M.

\textbf{Performance of static sentence embedding. }
Table~\ref{table:ablation_alignment} also reports the performance of static dense sentence embedding (number $4$) on the six datasets. The technique has achieved performance gains in 16 out of 18 metrics over the six datasets, with two performance drops of P@3 and P@5 on Eurlex-4K. On the three large-scale datasets: Wiki-500K, Amazon-670K and Amazon-3M, the performance gains in P@1 are $0.66 \%$, $0.39\%$ and $0.95\%$, respectively. There are three types of text features in our proposed MatchXML: sparse TF-IDF features, dense text features fine-tuned from pre-trained Transformer, and the static dense sentence embeddings extracted from Sentence-T5. The sparse TF-IDF features contains the global statistical information of input text, but it does not capture the semantic information. The dense text features fine-tuned from pre-trained Transformer are likely to lose parts of textual information due to the truncation operation (i.e., context window size of 512 tokens), while the static dense sentence embeddings can support much longer text sequence than the fine-tuned text embeddings from the encoder Transformer. Therefore, the static dense sentence embeddings can be considered as an effective complement to the sparse TF-IDF features and dense text features fine-tuned from pre-trained Transformer. As shown in Table~\ref{table:ablation_alignment} (number 4), including the static dense sentence embeddings boosts the performance of MatchXML consistently over the sparse TF-IDF baselines and the fine-tuned dense text feature baselines.

% \section{Limitations and Societal Impacts???}

\section{Conclusion}
\label{sec_conclustion}
This paper proposes MatchXML, a novel text-label matching framework, for the task of XMC. We introduce \emph{label2vec} to train the dense label embeddings to construct the Hierarchical Label Tree, where the dense label vectors have shown superior performance over the sparse TF-IDF label representations. In the fine-tuning stage of MatchXML, we formulate the multi-label text classification as the text-label matching problem within a mini-batch, leading to robust and effective dense text representations for XMC. In addition, we extract the static sentence embeddings from the pre-trained Sentence Transformer and incorporate them into our MatchXML to boost the performance further. Empirical study has demonstrated the superior performance of MatchXML in terms of classification accuracy and training speed over six benchmark datasets. It is worthy mentioning that although we propose MatchXML in the context of text classification, our framework is general and can be extended readily to other modalities for XMC, including image, audio, and video, etc. as long as a modality-specific encoder is available.

The training of MatchXML consists of four stages: training of \emph{label2vec}, construction of HLT, fine-tuning the text encoder, and training a linear classifier. As of future work, we plan to explore an end-to-end training approach to improve the performance of XMC further.

\bibliographystyle{IEEEtran}
\bibliography{main}

\end{document}